\documentclass[11pt]{article}
\pdfoutput=1
\usepackage[utf8]{inputenc} 
\usepackage{geometry}
\usepackage{graphicx}
\usepackage{array} 
\usepackage{verbatim} 
\usepackage{subfig}
\usepackage[]{natbib}
\usepackage{amsmath}
\usepackage{bbm}
\usepackage{stmaryrd}
\usepackage{hyperref}

\title{word2vec Parameter Learning Explained}
\author{Xin Rong \\ ronxin@umich.edu}
\date{}  % this makes date empty

\begin{document}
\maketitle

\begin{abstract}
The word2vec model and application by Mikolov et al. have attracted a great amount of attention in recent two years. The vector representations of words learned by word2vec models have been shown to carry semantic meanings and are useful in various NLP tasks. As an increasing number of researchers would like to experiment with word2vec or similar techniques, I notice that there lacks a material that comprehensively explains the parameter learning process of word embedding models in details, thus preventing researchers that are non-experts in neural networks from understanding the working mechanism of such models.

This note provides detailed derivations and explanations of the parameter update equations of the word2vec models, including the original continuous bag-of-word (CBOW) and skip-gram (SG) models, as well as advanced optimization techniques, including hierarchical softmax and negative sampling. Intuitive interpretations of the gradient equations are also provided alongside mathematical derivations.

In the appendix, a review on the basics of neuron networks and backpropagation is provided. I also created an interactive demo, wevi, to facilitate the intuitive understanding of the model.\footnote{An online interactive demo is available at: \url{http://bit.ly/wevi-online}.}
\end{abstract}

\section{Continuous Bag-of-Word Model}

\subsection{One-word context}
We start from the simplest version of the continuous bag-of-word model (CBOW) introduced in \cite{mikolov_efficient_2013}. We assume that there is only one word considered per context, which means the model will predict one target word given one context word, which is like a bigram model. For readers who are new to neural networks, it is recommended that one go through Appendix~\ref{sec:basics} for a quick review of the important concepts and terminologies before proceeding further.

\begin{figure}[htbp]
\centering
\includegraphics[width=4in]{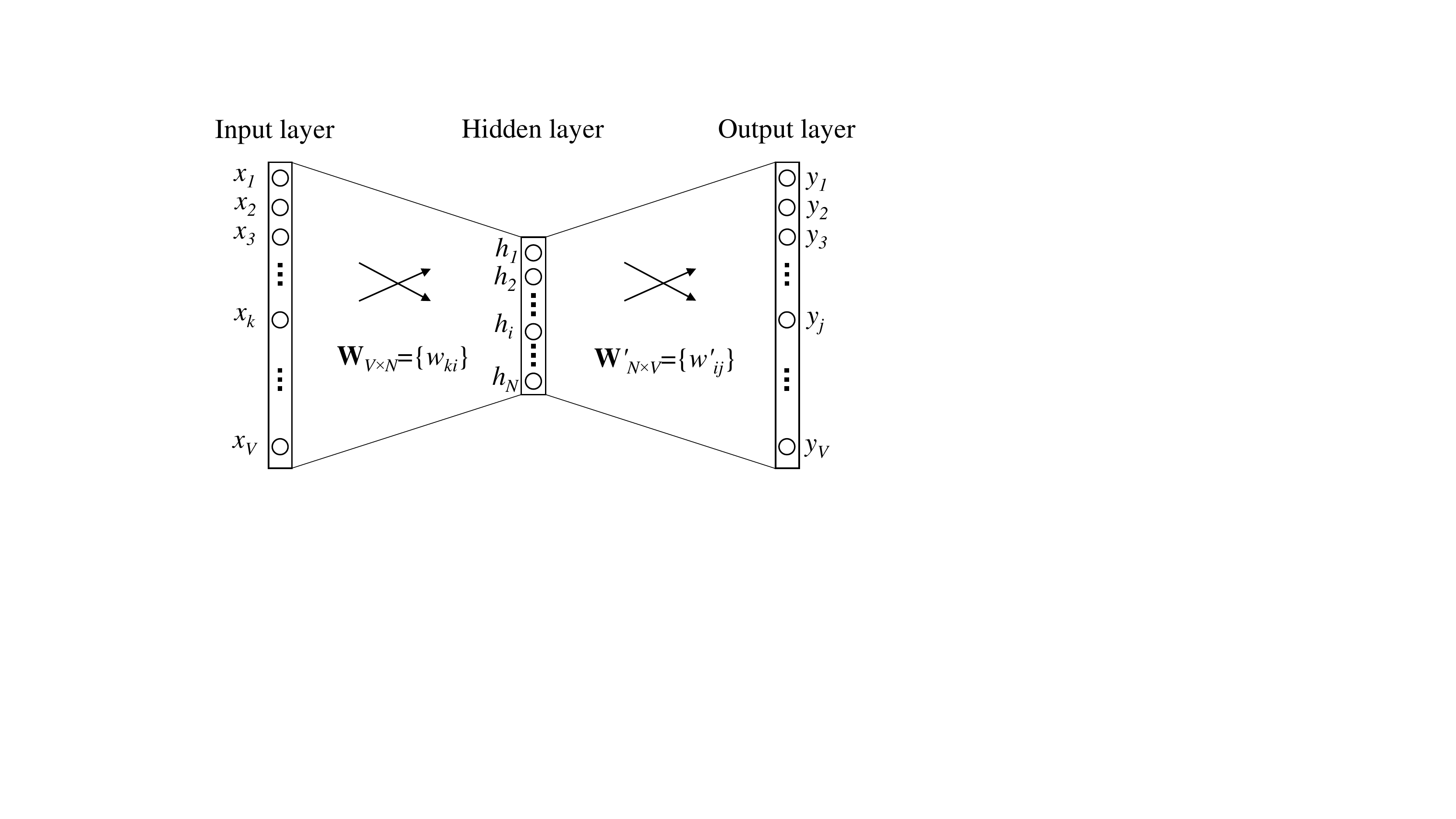}
\caption{A simple CBOW model with only one word in the context}
\label{fig:cbow-simple}
\end{figure}

Figure~\ref{fig:cbow-simple} shows the network model under the simplified context definition\footnote{In Figures~\ref{fig:cbow-simple}, \ref{fig:cbow}, \ref{fig:skip-gram}, and the rest of this note, $\mathbf{W}'$ is not the transpose of $\mathbf{W}$, but a different matrix instead.}. In our setting, the vocabulary size is $V$, and the hidden layer size is $N$. The units on adjacent layers are fully connected. The input is a one-hot encoded vector, which means for a given input context word, only one out of $V$ units, $\{x_1, \cdots, x_V\}$, will be 1, and all other units are 0.

The weights between the input layer and the output layer can be represented by a $V\times N$ matrix $\mathbf{W}$. Each row of $\mathbf{W}$ is the $N$-dimension vector representation $\mathbf{v}_w$ of the associated word of the input layer. Formally, row $i$ of $\textbf{W}$ is $\mathbf{v}_w^T$. Given a context (a word), assuming $x_k = 1$ and $x_{k'} = 0$ for $k' \neq k$, we have 
\begin{equation}
\mathbf{h} = \mathbf{W}^T \mathbf{x} = \mathbf{W}_{(k, \cdot)}^T := \mathbf{v}_{w_I}^T,
\label{eq:cbow-h}
\end{equation}
which is essentially copying the $k$-th row of $\mathbf{W}$ to $\mathbf{h}$. $\mathbf{v}_{w_I}$ is the vector representation of the input word $w_I$. This implies that the link (activation) function of the hidden layer units is simply \textit{linear} (i.e., directly passing its weighted sum of inputs to the next layer).

From the hidden layer to the output layer, there is a different weight matrix $\mathbf{W}' = \{w'_{ij}\}$, which is an $N\times V$ matrix. Using these weights, we can compute a score $u_j$ for each word in the vocabulary,
\begin{equation}
u_j = {\mathbf{v}'_{w_j}}^T \mathbf{h},
\label{eq:cbow-uj}
\end{equation}
where $\mathbf{v}'_{w_j}$ is the $j$-th column of the matrix $\mathbf{W'}$. Then we can use softmax, a log-linear classification model, to obtain the posterior distribution of words, which is a multinomial distribution.
\begin{equation}
p(w_j | w_I) = y_j = \frac{\exp(u_j)}{\sum_{j'=1}^V\exp(u_{j'})},
\label{eq:cbow-yj}
\end{equation}
where $y_j$ is the output of the $j$-the unit in the output layer. Substituting (\ref{eq:cbow-h}) and (\ref{eq:cbow-uj}) into (\ref{eq:cbow-yj}), we obtain
\begin{equation}
p(w_j | w_I) = \frac{\exp\left({\mathbf{v}'_{w_j}}^T\mathbf{v}_{w_I}\right)}{\sum_{j'=1}^V\exp\left({\mathbf{v}'_{w_{j'}}}^T\mathbf{v}_{w_I}\right)}
\label{eq:cbow-pwo}
\end{equation}

Note that $\mathbf{v}_w$ and $\mathbf{v}'_w$ are two representations of the word $w$. $\mathbf{v}_w$ comes from rows of $\mathbf{W}$, which is the input$\rightarrow$hidden weight matrix, and $\mathbf{v}'_w$ comes from columns of $\mathbf{W'}$, which is the hidden$\rightarrow$output matrix. In subsequent analysis, we call $\mathbf{v}_w$ as the ``\textbf{input vector}'', and $\mathbf{v}'_w$ as the ``\textbf{output vector}'' of the word $w$.

\subsubsection*{Update equation for hidden$\rightarrow$output weights}

Let us now derive the weight update equation for this model. Although the actual computation is impractical (explained below), we are doing the derivation to gain insights on this original model with no tricks applied. For a review of basics of backpropagation, see Appendix~\ref{sec:basics}.

The training objective (for one training sample) is to maximize (\ref{eq:cbow-pwo}), the conditional probability of observing the actual output word $w_O$ (denote its index in the output layer as $j^*$) given the input context word $w_I$ with regard to the weights.
\begin{eqnarray}
\max p(w_O | w_I)
&=& \max y_{j^*} \\
&=& \max \log y_{j^*} \\
&=& u_{j^*} - \log\sum_{j'=1}^V\exp(u_{j'}) := -E,
\label{eq:cbow-simple-objective}
\end{eqnarray}
where $E=-\log p(w_O|w_I)$ is our loss function (we want to minimize $E$), and $j^*$ is the index of the actual output word in the output layer. Note that this loss function can be understood as a special case of the cross-entropy measurement between two probabilistic distributions.

Let us now derive the update equation of the weights between hidden and output layers. Take the derivative of $E$ with regard to $j$-th unit's net input $u_j$, we obtain
\begin{equation}
\frac{\partial E}{\partial u_j} = y_j - t_j := e_j
\label{eq:cbow-eui}
\end{equation}
where $t_j = \mathbbm{1}(j = j^*)$, i.e., $t_j$ will only be 1 when the $j$-th unit is the actual output word, otherwise $t_j = 0$.  Note that this derivative is simply the prediction error $e_j$ of the output layer.

Next we take the derivative on $w'_{ij}$ to obtain the gradient on the hidden$\rightarrow$output weights.
\begin{equation}
\frac{\partial E}{\partial w'_{ij}} 
= \frac{\partial E}{\partial u_j} \cdot \frac{\partial u_j}{\partial w'_{ij}}
= e_j \cdot h_i
\end{equation}
Therefore, using stochastic gradient descent, we obtain the weight updating equation for hidden$\rightarrow$output weights:
\begin{equation}
{w'_{ij}}^\text{(new)} = {w'_{ij}}^\text{(old)} - \eta\cdot e_j \cdot h_i.
\end{equation}
or
\begin{equation}
{\mathbf{v}'_{w_j}}^\text{(new)} = {\mathbf{v}'_{w_j}}^\text{(old)} - \eta\cdot e_j\cdot \mathbf{h} \hspace{2.5em}\text{for }j=1,2,\cdots,V.
\label{eq:cbow-update-wij}
\end{equation}
where $\eta > 0$ is the learning rate, $e_j = y_j - t_j$, and $h_i$ is the $i$-th unit in the hidden layer; $\mathbf{v}'_{w_j}$ is the output vector of $w_j$. Note that this update equation implies that we have to go through every possible word in the vocabulary, check its output probability $y_j$, and compare $y_j$ with its expected output $t_j$ (either 0 or 1). If $y_j > t_j$ (``overestimating''), then we subtract a proportion of the hidden vector $\mathbf{h}$ (i.e., $\mathbf{v}_{w_I}$) from $\mathbf{v}'_{w_j}$, thus making $\mathbf{v}'_{w_j}$ farther away from $\mathbf{v}_{w_I}$; if $y_j < t_j$ (``underestimating'', which is true only if $t_j=1$, i.e., $w_j=w_O$), we add some $\mathbf{h}$ to $\mathbf{v}'_{w_O}$, thus making $\mathbf{v}'_{w_O}$ closer\footnote{Here when I say ``closer'' or ``farther'', I meant using the inner product instead of Euclidean as the distance measurement.} to $\mathbf{v}_{w_I}$. If $y_j$ is very close to $t_j$, then according to the update equation, very little change will be made to the weights. Note, again, that $\mathbf{v}_w$ (input vector) and $\mathbf{v}'_w$ (output vector) are two different vector representations of the word $w$.

\subsubsection*{Update equation for input$\rightarrow$hidden weights}

Having obtained the update equations for $\mathbf{W'}$, we can now move on to $\mathbf{W}$. We take the derivative of $E$ on the output of the hidden layer, obtaining
\begin{equation}
\frac{\partial E}{\partial h_i} 
= \sum_{j=1}^V \frac{\partial E}{\partial u_j} \cdot \frac{\partial u_j}{\partial h_i} = \sum_{j=1}^V e_j \cdot w'_{ij} := \text{EH}_i
\label{eq:cbow-ehi}
\end{equation}
where $h_i$ is the output of the $i$-th unit of the hidden layer; $u_j$ is defined in (\ref{eq:cbow-uj}), the net input of the $j$-th unit in the output layer; and $e_j = y_j - t_j$ is the prediction error of the $j$-th word in the output layer. $\text{EH}$, an $N$-dim vector, is the sum of the output vectors of all words in the vocabulary, weighted by their prediction error.

Next we should take the derivative of $E$ on $\mathbf{W}$. First, recall that the hidden layer performs a linear computation on the values from the input layer. Expanding the vector notation in (\ref{eq:cbow-h}) we get
\begin{equation}
h_i = \sum_{k=1}^V x_k\cdot w_{ki}
\end{equation}
Now we can take the derivative of $E$ with regard to each element of $\mathbf{W}$, obtaining
\begin{equation}
\frac{\partial E}{\partial w_{ki}}
= \frac{\partial E}{\partial h_i} \cdot \frac{\partial h_i}{\partial w_{ki}}
= \text{EH}_i \cdot x_k
\end{equation}
This is equivalent to the tensor product of $\mathbf{x}$ and $\text{EH}$, i.e.,
\begin{equation}
\frac{\partial E}{\partial \mathbf{W}} = \mathbf{x} \otimes \text{EH} = \mathbf{x} \text{EH}^T
\end{equation}
from which we obtain a $V\times N$ matrix. Since only one component of $\mathbf{x}$ is non-zero, only one row of $\frac{\partial E}{\partial \mathbf{W}}$ is non-zero, and the value of that row is $\text{EH}^T$, an $N$-dim vector. We obtain the update equation of $\mathbf{W}$ as
\begin{equation}
\mathbf{v}_{w_I}^\text{(new)} = \mathbf{v}_{w_I}^\text{(old)} - \eta\text{EH}^T
\label{eq:cbow-update-wki}
\end{equation}
where $\mathbf{v}_{w_I}$ is a row of $\mathbf{W}$, the ``input vector'' of the only context word, and is the only row of $\mathbf{W}$ whose derivative is non-zero. All the other rows of $\mathbf{W}$ will remain unchanged after this iteration, because their derivatives are zero.

Intuitively, since vector $\text{EH}$ is the sum of output vectors of all words in vocabulary weighted by their prediction error $e_j = y_j- t_j$, we can understand (\ref{eq:cbow-update-wki}) as adding a portion of every output vector in vocabulary to the input vector of the context word. If, in the output layer, the probability of a word $w_j$ being the output word is overestimated ($y_j > t_j$), then the input vector of the context word $w_I$ will tend to move farther away from the output vector of $w_j$; conversely if the probability of $w_j$ being the output word is underestimated ($y_j < t_j$), then the input vector $w_I$ will tend to move closer to the output vector of $w_j$; if the probability of $w_j$ is fairly accurately predicted, then it will have little effect on the movement of the input vector of $w_I$. The movement of the input vector of $w_I$ is determined by the prediction error of all vectors in the vocabulary; the larger the prediction error, the more significant effects a word will exert on the movement on the input vector of the context word. 

As we iteratively update the model parameters by going through context-target word pairs generated from a training corpus, the effects on the vectors will accumulate. We can imagine that the output vector of a word $w$ is ``dragged'' back-and-forth by the input vectors of $w$'s co-occurring neighbors, as if there are physical strings between the vector of $w$ and the vectors of its neighbors. Similarly, an input vector can also be considered as being dragged by many output vectors. This interpretation can remind us of gravity, or force-directed graph layout. The equilibrium length of each imaginary string is related to the strength of cooccurrence between the associated pair of words, as well as the learning rate. After many iterations, the relative positions of the input and output vectors will eventually stabilize.
%This may be the explanation of why this model may obtain ``king'' - ``queen'' = ``man'' - ``woman''. Imagine the word ``king'' being dragged around by different forces from the words it intimately co-occurs with; it may end up at a stabilized position determined by its most frequently co-occurring words. Because we see these word pairs so many times, these top-frequent neighbors will dominate the movements of the target word.

\subsection{Multi-word context}

\begin{figure}[htb]
\centering
\includegraphics[width=3in]{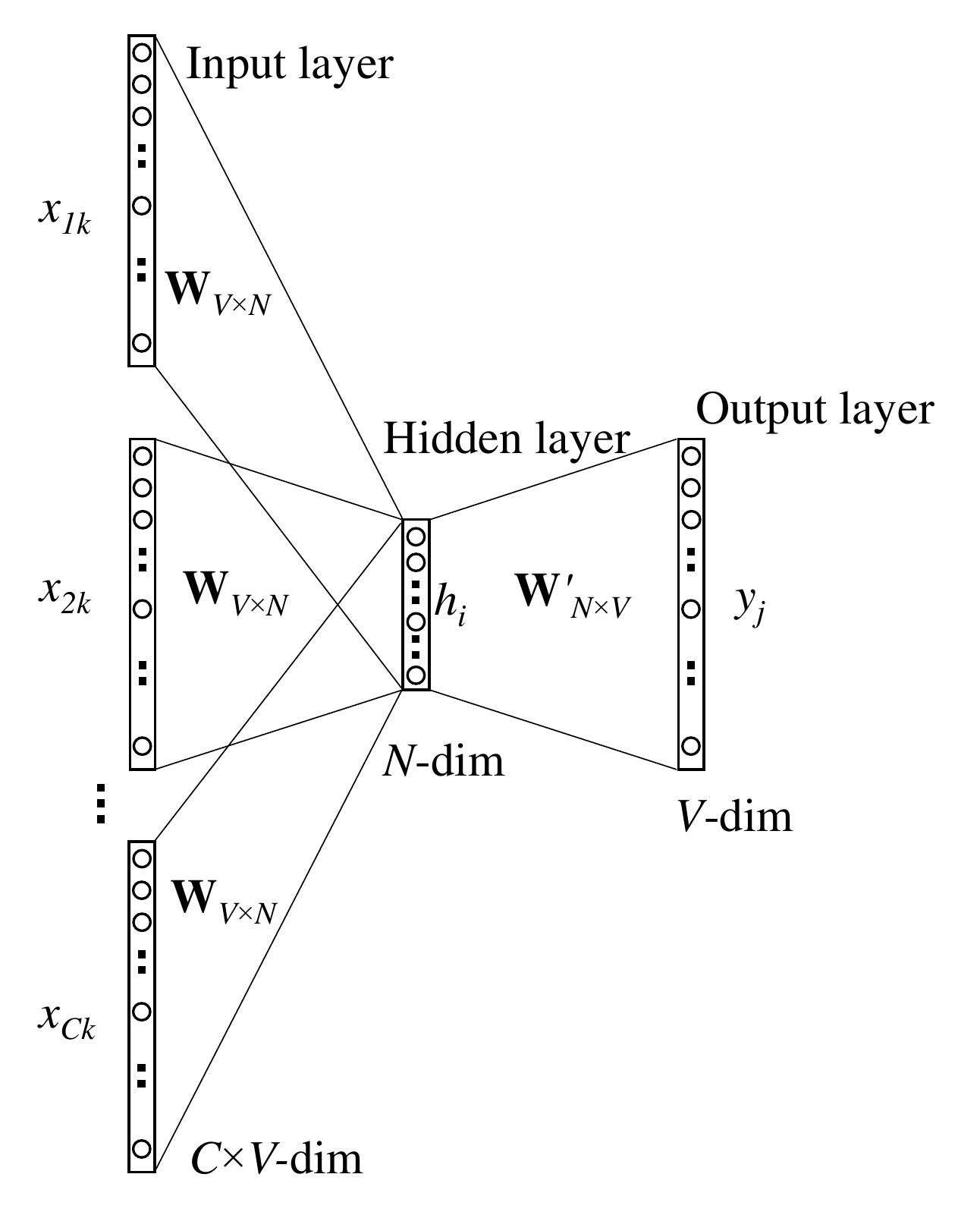}
\caption{Continuous bag-of-word model}
\label{fig:cbow}
\end{figure}

Figure~\ref{fig:cbow} shows the CBOW model with a multi-word context setting.
When computing the hidden layer output, instead of directly copying the input vector of the input context word, the CBOW model takes the average of the vectors of the input context words, and use the product of the input$\rightarrow$hidden weight matrix and the average vector as the output.
\begin{eqnarray}
\mathbf{h} &=& \frac{1}{C}\mathbf{W}^T(\mathbf{x}_1+\mathbf{x}_2+\cdots+\mathbf{x}_C) \\
&=& \frac{1}{C}(\mathbf{v}_{w_1} + \mathbf{v}_{w_2} + \cdots + \mathbf{v}_{w_C})^T
\label{eq:cbow-complex-h}
\end{eqnarray}
where $C$ is the number of words in the context, $w_1, \cdots, w_C$ are the words the in the context, and $\mathbf{v}_w$ is the input vector of a word $w$. The loss function is
\begin{eqnarray}
E &=& = -\log p(w_O | w_{I,1}, \cdots, w_{I,C}) \\
&=& -u_{j^*} + \log\sum_{j'=1}^V\exp(u_{j'}) \\
&=& -{\mathbf{v}'_{w_O}}^T\cdot \mathbf{h} + \log\sum_{j'=1}^V\exp({\mathbf{v}'_{w_j}}^T\cdot \mathbf{h})
\end{eqnarray}
which is the same as (\ref{eq:cbow-simple-objective}), the objective of the one-word-context model, except that $\mathbf{h}$ is different, as defined in (\ref{eq:cbow-complex-h}) instead of (\ref{eq:cbow-h}).

The update equation for the hidden$\rightarrow$output weights stay the same as that for the one-word-context model (\ref{eq:cbow-update-wij}). We copy it here:
\begin{equation}
{\mathbf{v}'_{w_j}}^\text{(new)} = {\mathbf{v}'_{w_j}}^\text{(old)} - \eta\cdot e_j\cdot \mathbf{h}  \hspace{2.5em}\text{for }j=1,2,\cdots,V.
\label{eq:cbow-complex-update-wij}
\end{equation}
Note that we need to apply this to every element of the hidden$\rightarrow$output weight matrix for each training instance.

The update equation for input$\rightarrow$hidden weights is similar to (\ref{eq:cbow-update-wki}), except that now we need to apply the following equation for every word $w_{I,c}$ in the context:
\begin{equation}
\mathbf{v}_{w_{I,c}}^\text{(new)} = \mathbf{v}_{w_{I,c}}^\text{(old)} - \frac{1}{C}\cdot\eta\cdot\text{EH}^T  \hspace{2.5em}\text{for }c=1,2,\cdots,C.
\label{eq:cbow-complex-update-wki}
\end{equation}
where $\mathbf{v}_{w_{I,c}}$ is the input vector of the $c$-th word in the input context; $\eta$ is a positive learning rate; and $\text{EH} = \frac{\partial E}{\partial h_i}$ is given by (\ref{eq:cbow-ehi}). The intuitive understanding of this update equation is the same as that for (\ref{eq:cbow-update-wki}).

\section{Skip-Gram Model}

\begin{figure}[htbp]
\centering
\includegraphics[width=3in]{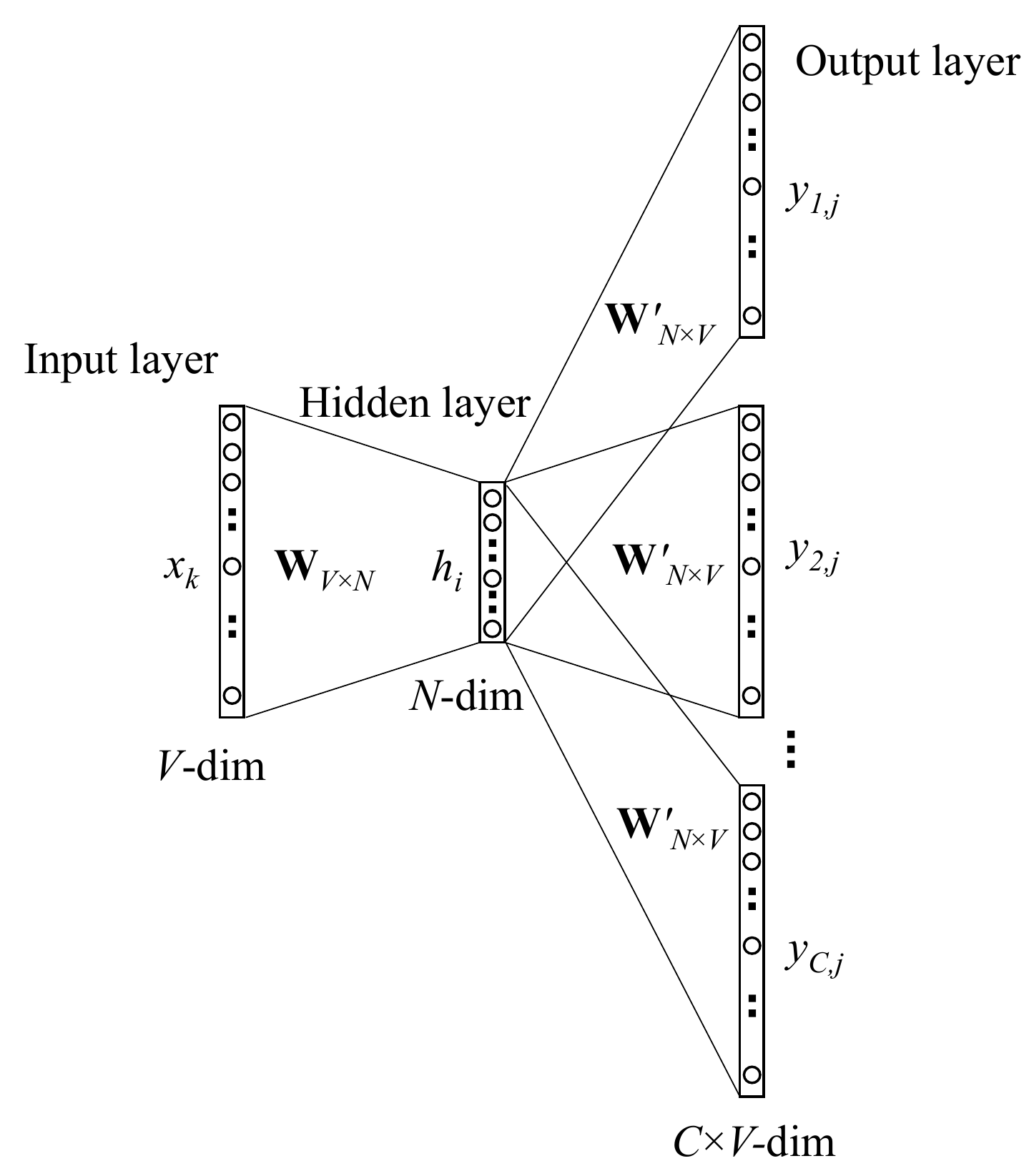}
\caption{The skip-gram model.}
\label{fig:skip-gram}
\end{figure}

The skip-gram model is introduced in \cite{mikolov_efficient_2013,mikolov_distributed_2013}. Figure~\ref{fig:skip-gram} shows the skip-gram model. It is the opposite of the CBOW model. The target word is now at the input layer, and the context words are on the output layer. 

We still use $\mathbf{v}_{w_I}$ to denote the input vector of the only word on the input layer, and thus we have the same definition of the hidden-layer outputs $\mathbf{h}$ as in (\ref{eq:cbow-h}), which means $\mathbf{h}$ is simply copying (and transposing) a row of the input$\rightarrow$hidden weight matrix, $\mathbf{W}$, associated with the input word $w_I$. We copy the definition of $\mathbf{h}$ below:
\begin{equation}
\mathbf{h} = \mathbf{W}_{(k, \cdot)}^T := \mathbf{v}_{w_I}^T,
\end{equation}

On the output layer, instead of outputing one multinomial distribution, we are outputing $C$ multinomial distributions. Each output is computed using the same hidden$\rightarrow$output matrix:
\begin{equation}
p(w_{c,j} = w_{O,c} | w_I) = y_{c,j} = \frac{\exp(u_{c,j})}{\sum_{j'=1}^V\exp(u_{j'})}
\end{equation}
where $w_{c,j}$ is the $j$-th word on the $c$-th panel of the output layer; $w_{O,c}$ is the actual $c$-th word in the output context words; $w_I$ is the only input word; $y_{c,j}$ is the output of the $j$-th unit on the $c$-th panel of the output layer; $u_{c,j}$ is the net input of the $j$-th unit on the $c$-th panel of the output layer. Because the output layer panels share the same weights, thus
\begin{equation}
u_{c,j} = u_j = {\mathbf{v}'_{w_j}}^T\cdot\mathbf{h}, \text{ for } c = 1, 2, \cdots, C
\end{equation}
where ${\mathbf{v}'_{w_j}}$ is the output vector of the $j$-th word in the vocabulary, $w_j$, and also ${\mathbf{v}'_{w_j}}$ is taken from a column of the hidden$\rightarrow$output weight matrix, $\mathbf{W}'$.

The derivation of parameter update equations is not so different from the one-word-context model. The loss function is changed to
\begin{eqnarray}
E &=& -\log p(w_{O,1}, w_{O,2}, \cdots, w_{O,C} | w_I) \\
&=& -\log \prod_{c=1}^C\frac{\exp(u_{c,j_c^*})}{\sum_{j'=1}^V\exp(u_{j'})} \\
&=& -\sum_{c=1}^C u_{j_c^*} + C\cdot\log\sum_{j'=1}^V\exp(u_{j'})
\end{eqnarray}
where $j_c^*$ is the index of the actual $c$-th output context word in the vocabulary.

We take the derivative of $E$ with regard to the net input of every unit on every panel of the output layer, $u_{c,j}$ and obtain
\begin{equation}
\frac{\partial E}{\partial u_{c,j}} = y_{c,j} - t_{c,j} := e_{c,j}
\end{equation}
which is the prediction error on the unit, the same as in (\ref{eq:cbow-eui}). For notation simplicity, we define a $V$-dimensional vector $\text{EI} = \{\text{EI}_1, \cdots, \text{EI}_V\}$ as the sum of prediction errors over all context words:
\begin{equation}
\text{EI}_j = \sum_{c=1}^C e_{c,j}
\end{equation}
Next, we take the derivative of $E$ with regard to the hidden$\rightarrow$output matrix $\mathbf{W}'$, and obtain
\begin{equation}
\frac{\partial E}{\partial w'_{ij}} 
= \sum_{c=1}^C\frac{\partial E}{\partial u_{c,j}} \cdot \frac{\partial u_{c,j}}{\partial w'_{ij}}
= \text{EI}_j\cdot h_i
\end{equation}
Thus we obtain the update equation for the hidden$\rightarrow$output matrix $\mathbf{W}'$,
\begin{equation}
{w'_{ij}}^\text{(new)} = {w'_{ij}}^\text{(old)} - \eta \cdot \text{EI}_j\cdot h_i
\label{eq:skipgram-update-wij}
\end{equation}
or
\begin{equation}
{\mathbf{v}'_{w_j}}^\text{(new)} = {\mathbf{v}'_{w_j}}^\text{(old)} - \eta\cdot\text{EI}_j\cdot\mathbf{h} \hspace{2.5em}\text{for }j=1,2,\cdots,V.
\end{equation}
The intuitive understanding of this update equation is the same as that for (\ref{eq:cbow-update-wij}), except that the prediction error is summed across all context words in the output layer. Note that we need to apply this update equation for every element of the hidden$\rightarrow$output matrix for each training instance.

The derivation of the update equation for the input$\rightarrow$hidden matrix is identical to (\ref{eq:cbow-ehi}) to (\ref{eq:cbow-update-wki}), except taking into account that the prediction error $e_j$ is replaced with $\text{EI}_j$. We directly give the update equation:
\begin{equation}
\textbf{v}_{w_I}^\text{(new)} = \textbf{v}_{w_I}^\text{(old)} - \eta\cdot\text{EH}^T
\label{eq:skipgram-update-wki}
\end{equation}
where \text{EH} is an $N$-dim vector, each component of which is defined as
\begin{equation}
\text{EH}_i = \sum_{j=1}^V\text{EI}_j\cdot w'_{ij}.
\end{equation}
The intuitive understanding of (\ref{eq:skipgram-update-wki}) is the same as that for (\ref{eq:cbow-update-wki}).

\section{Optimizing Computational Efficiency}
So far the models we have discussed (``bigram'' model, CBOW and skip-gram) are both in their original forms, without any efficiency optimization tricks being applied. 

For all these models, there exist two vector representations for each word in the vocabulary: the input vector $\mathbf{v}_w$, and the output vector $\mathbf{v}'_w$. Learning the input vectors is cheap; but learning the output vectors is very expensive. From the update equations (\ref{eq:cbow-complex-update-wij}) and (\ref{eq:skipgram-update-wij}), we can find that, in order to update $\mathbf{v}'_w$, for each training instance, we have to iterate through every word $w_j$ in the vocabulary, compute their net input $u_j$, probability prediction $y_j$ (or $y_{c,j}$ for skip-gram), their prediction error $e_j$ (or $\text{EI}_j$ for skip-gram), and finally use their prediction error to update their output vector $\mathbf{v}'_j$. 

Doing such computations for all words for every training instance is very expensive, making it impractical to scale up to large vocabularies or large training corpora. To solve this problem, an intuition is to limit the number of output vectors that must be updated per training instance. One elegant approach to achieving this is hierarchical softmax; another approach is through sampling, which will be discussed in the next section. 

Both tricks optimize only the computation of the updates for output vectors. In our derivations, we care about three values: (1) $E$, the new objective function; (2) $\frac{\partial E}{\partial \mathbf{v}'_{w}}$, the new update equation for the output vectors; and (3) $\frac{\partial E}{\partial \mathbf{h}}$, the weighted sum of predictions errors to be backpropagated for updating input vectors.

\subsection{Hierarchical Softmax}
\begin{figure}[htbp]
\centering
\includegraphics[width=3in]{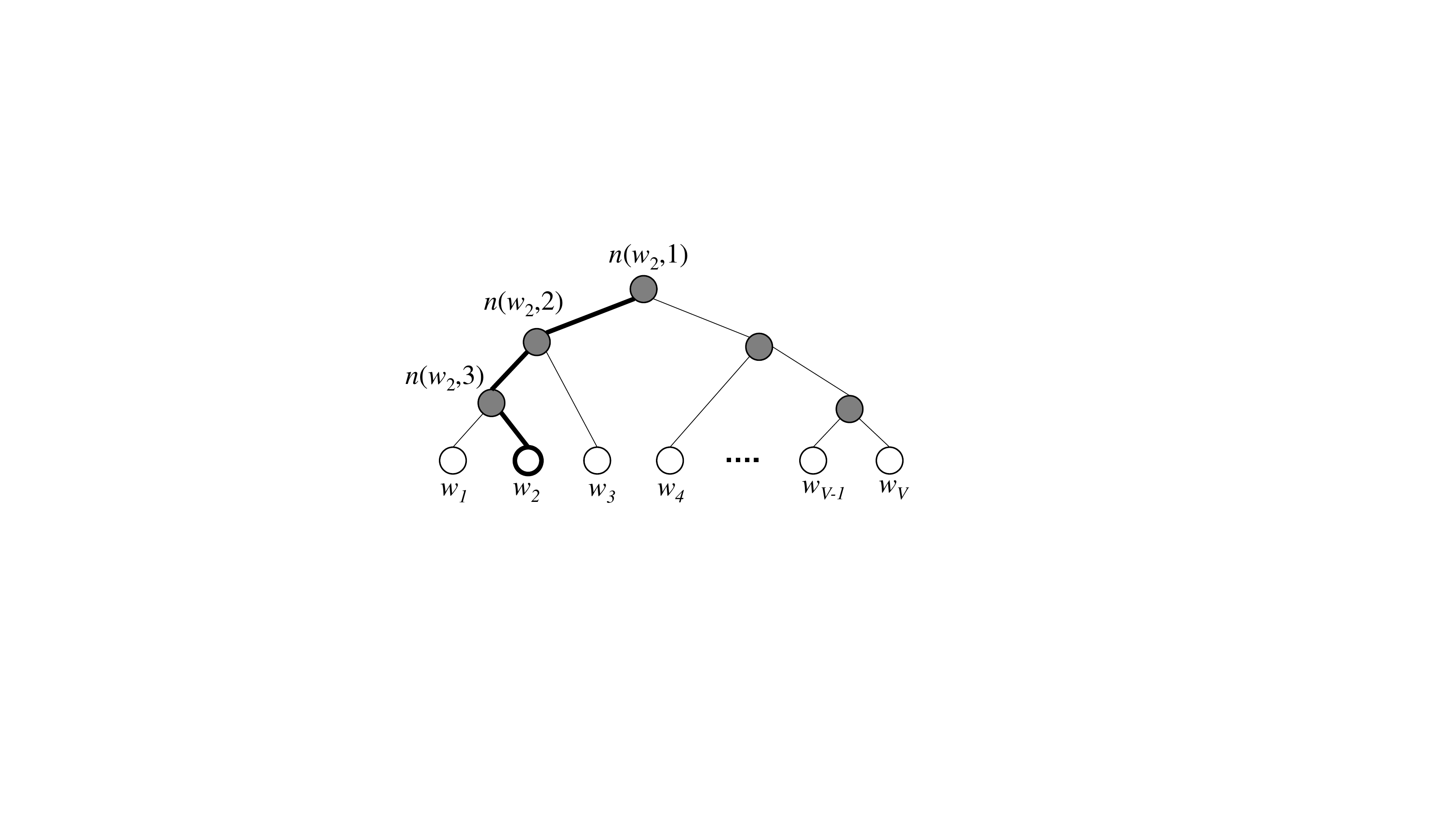}
\caption{An example binary tree for the hierarchical softmax model. The white units are words in the vocabulary, and the dark units are inner units. An example path from root to $w_2$ is highlighted. In the example shown, the length of the path $L(w_2) = 4$. $n(w,j)$ means the $j$-th unit on the path from root to the word $w$.}
\label{fig:hsoftmax}
\end{figure}

Hierarchical softmax is an efficient way of computing softmax \citep{morin_hierarchical_2005,mnih_scalable_2009}. The model uses a binary tree to represent all words in the vocabulary. The $V$ words must be leaf units of the tree. It can be proved that there are $V-1$ inner units. For each leaf unit, there exists a unique path from the root to the unit; and this path is used to estimate the probability of the word represented by the leaf unit. See Figure~\ref{fig:hsoftmax} for an example tree.

In the hierarchical softmax model, there is no output vector representation for words. Instead, each of the $V-1$ inner units has an output vector $\mathbf{v}'_{n(w,j)}$. And the probability of a word being the output word is defined as
\begin{equation}
p(w = w_O) = \prod_{j=1}^{L(w)-1}\sigma\left(\llbracket n(w,j+1)=\text{ch}(n(w,j))\rrbracket\cdot {\mathbf{v}'_{n(w,j)}}^T \mathbf{h}\right)
\label{eq:hs}
\end{equation}
where $\text{ch}(n)$ is the left child of unit $n$; $\mathbf{v}'_{n(w,j)}$ is the vector representation (``output vector'') of the inner unit $n(w,j)$; $\mathbf{h}$ is the output value of the hidden layer (in the skip-gram model $\mathbf{h} = \mathbf{v}_{w_I}$; and in CBOW, $\mathbf{h}=\frac{1}{C}\sum_{c=1}^C\mathbf{v}_{w_c}$); $\llbracket x\rrbracket$ is a special function defined as
\begin{equation}
\llbracket x\rrbracket = \begin{cases}1 & \text{if }x \text{ is true;}\\-1 & \text{otherwise.}\end{cases}
\end{equation}

Let us intuitively understand the equation by going through an example. Looking at Figure~\ref{fig:hsoftmax}, suppose we want to compute the probability that $w_2$ being the output word. We define this probability as the probability of a random walk starting from the root ending at the leaf unit in question. At each inner unit (including the root unit), we need to assign the probabilities of going left and going right.\footnote{While an inner unit of a binary tree may not always have both children, a binary Huffman tree's inner units always do. Although theoretically one can use many different types of trees for hierarchical softmax, word2vec uses a binary Huffman tree for fast training.}
We define the probability of going left at an inner unit $n$ to be
\begin{equation}
p(n,\text{left}) = \sigma\left({\mathbf{v}'_n}^T\cdot\mathbf{h}\right)
\end{equation}
which is determined by both the vector representation of the inner unit, and the hidden layer's output value (which is then determined by the vector representation of the input word(s)). Apparently the probability of going right at unit $n$ is
\begin{equation}
p(n,\text{right})
= 1 - \sigma\left({\mathbf{v}'_n}^T\cdot\mathbf{h}\right)
= \sigma\left(-{\mathbf{v}'_n}^T\cdot\mathbf{h}\right)
\end{equation}
Following the path from the root to $w_2$ in Figure~\ref{fig:hsoftmax}, we can compute the probability of $w_2$ being the output word as
\begin{eqnarray}
p(w_2 = w_O) 
&=& 
	p\left(n(w_2,1),\text{left}\right)
	\cdot p\left(n(w_2,2),\text{left}\right)
	\cdot p\left(n(w_2,3),\text{right}\right) \\
&=& 
	\sigma\left({\mathbf{v}'_{n(w_2,1)}}^T\mathbf{h}\right)
	\cdot\sigma\left({\mathbf{v}'_{n(w_2,2)}}^T\mathbf{h}\right)
	\cdot\sigma\left(-{\mathbf{v}'_{n(w_2,3)}}^T\mathbf{h}\right)
\end{eqnarray}
which is exactly what is given by (\ref{eq:hs}). It should not be hard to verify that 
\begin{equation}
\sum_{i=1}^V p(w_i=w_O) = 1
\end{equation}
making the hierarchical softmax a well defined multinomial distribution among all words.

Now let us derive the parameter update equation for the vector representations of the inner units. For simplicity, we look at the one-word context model first. Extending the update equations to CBOW and skip-gram models is easy. 

For the simplicity of notation, we define the following shortenizations without introducing ambiguity:
\begin{equation}
\llbracket\cdot\rrbracket := \llbracket n(w,j+1)=\text{ch}(n(w,j))\rrbracket
\end{equation}
\begin{equation}
\mathbf{v}'_j := \mathbf{v}'_{n_{w,j}}
\end{equation}

For a training instance, the error function is defined as
\begin{equation}
E = -\log p(w=w_O|w_I) = -\sum_{j=1}^{L(w)-1}\log\sigma(\llbracket\cdot\rrbracket{\mathbf{v}'_j}^T\mathbf{h})
\end{equation}

We take the derivative of $E$ with regard to $\mathbf{v}'_j\mathbf{h}$, obtaining
\begin{eqnarray}
\frac{\partial E}{\partial \mathbf{v}'_j\mathbf{h}}
&=& \left(
      \sigma(\llbracket\cdot\rrbracket{\mathbf{v}'_j}^T\mathbf{h}) - 1
    \right)
    \llbracket\cdot\rrbracket \\
&=& \begin{cases}
	\sigma({\mathbf{v}'_j}^T\mathbf{h}) - 1 & (\llbracket\cdot\rrbracket = 1) \\
	\sigma({\mathbf{v}'_j}^T\mathbf{h})& (\llbracket\cdot\rrbracket = -1)
	\end{cases} \\
&=& \sigma({\mathbf{v}'_j}^T\mathbf{h}) - t_j
\end{eqnarray}
where $t_j = 1$ if $\llbracket\cdot\rrbracket = 1$ and $t_j = 0$ otherwise.

Next we take the derivative of $E$ with regard to the vector representation of the inner unit $n(w,j)$ and obtain
\begin{equation}
\frac{\partial E}{\partial \mathbf{v}'_j}
= \frac{\partial E}{\partial \mathbf{v}'_j\mathbf{h}} \cdot \frac{\partial \mathbf{v}'_j\mathbf{h}}{\partial \mathbf{v}'_j}
= \left(\sigma({\mathbf{v}'_j}^T\mathbf{h}) - t_j\right)\cdot \mathbf{h}
\end{equation}
which results in the following update equation:
\begin{equation}
{\mathbf{v}'_j}^\text{(new)} = {\mathbf{v}'_j}^\text{(old)} - \eta\left(\sigma({\mathbf{v}'_j}^T\mathbf{h}) - t_j\right)\cdot \mathbf{h}
\end{equation}
which should be applied to $j=1, 2, \cdots, L(w)-1$. We can understand $\sigma({\mathbf{v}'_j}^T\mathbf{h}) - t_j$ as the prediction error for the inner unit $n(w,j)$. The ``task'' for each inner unit is to predict whether it should follow the left child or the right child in the random walk. $t_j=1$ means the ground truth is to follow the left child; $t_j=0$ means it should follow the right child. $\sigma({\mathbf{v}'_j}^T\mathbf{h})$ is the prediction result. For a training instance, if the prediction of the inner unit is very close to the ground truth, then its vector representation $\mathbf{v}'_j$ will move very little; otherwise $\mathbf{v}'_j$ will move in an appropriate direction by moving (either closer or farther away\footnote{Again, the distance measurement is inner product.} from $\mathbf{h}$) so as to reduce the prediction error for this instance. This update equation can be used for both CBOW and the skip-gram model. When used for the skip-gram model, we need to repeat this update procedure for each of the $C$ words in the output context.

In order to backpropagate the error to learn input$\rightarrow$hidden weights, we take the derivative of $E$ with regard to the output of the hidden layer and obtain
\begin{eqnarray}
\frac{\partial E}{\partial \mathbf{h}}
&=& \sum_{j=1}^{L(w)-1}\frac{\partial E}{\partial \mathbf{v}'_j\mathbf{h}} \cdot \frac{\partial \mathbf{v}'_j\mathbf{h}}{\partial \mathbf{h}} \\
&=& \sum_{j=1}^{L(w)-1}\left(\sigma({\mathbf{v}'_j}^T\mathbf{h}) - t_j\right)\cdot \mathbf{v}'_j \\
&:=& \text{EH}
\end{eqnarray}
which can be directly substituted into (\ref{eq:cbow-complex-update-wki}) to obtain the update equation for the input vectors of CBOW. For the skip-gram model, we need to calculate a $\text{EH}$ value for each word in the skip-gram context, and plug the sum of the $\text{EH}$ values into (\ref{eq:skipgram-update-wki}) to obtain the update equation for the input vector.

From the update equations, we can see that the computational complexity per training instance per context word is reduced from $O(V)$ to $O(\log(V))$, which is a big improvement in speed. We still have roughly the same number of parameters ($V-1$ vectors for inner-units compared to originally $V$ output vectors for words).

\subsection{Negative Sampling}
The idea of negative sampling is more straightforward than hierarchical softmax: in order to deal with the difficulty of having too many output vectors that need to be updated per iteration, we only update a sample of them.

Apparently the output word (i.e., the ground truth, or positive sample) should be kept in our sample and gets updated, and we need to sample a few words as negative samples (hence ``negative sampling''). A probabilistic distribution is needed for the sampling process, and it can be arbitrarily chosen. We call this distribution the noise distribution, and denote it as $P_n(w)$. One can determine a good distribution empirically.\footnote{As described in \citep{mikolov_distributed_2013}, word2vec uses a unigram distribution raised to the $\frac{3}{4}$th power for the best quality of results.}

In word2vec, instead of using a form of negative sampling that produces a well-defined posterior multinomial distribution, the authors argue that the following simplified training objective is capable of producing high-quality word embeddings:\footnote{\cite{goldberg_word2vec_2014} provide a theoretical analysis on the reason of using this objective function.}
\begin{equation}
E = -\log\sigma({\mathbf{v}'_{w_O}}^T\mathbf{h})
- \sum_{w_j\in\mathcal{W}_{\text{neg}}}\log\sigma(-{\mathbf{v}'_{w_j}}^T\mathbf{h})
\end{equation}
where $w_O$ is the output word (i.e., the positive sample), and
$\mathbf{v}'_{w_O}$ is its output vector; 
$\mathbf{h}$ is the output value of the hidden layer: $\mathbf{h}=\frac{1}{C}\sum_{c=1}^C\mathbf{v}_{w_c}$ in the CBOW model and  $\mathbf{h} = \mathbf{v}_{w_I}$ in the skip-gram model; $\mathcal{W}_{\text{neg}} = \{w_j | j = 1, \cdots, K\}$ is the set of words that are sampled based on $P_n(w)$, i.e., negative samples.

To obtain the update equations of the word vectors under negative sampling, we first take the derivative of $E$ with regard to the net input of the output unit $w_j$:
\begin{eqnarray}
\frac{\partial E}{\partial {\mathbf{v}'_{w_j}}^T\mathbf{h}} 
&=& \begin{cases} \sigma({\mathbf{v}'_{w_j}}^T\mathbf{h})-1  & \text{if }w_j = w_O\\
    \sigma({\mathbf{v}'_{w_j}}^T\mathbf{h}) & \text{if }w_j \in \mathcal{W}_{\text{neg}} \end{cases} \\
&=& \sigma({\mathbf{v}'_{w_j}}^T\mathbf{h}) - t_j
\end{eqnarray}
where $t_j$ is the ``label'' of word $w_j$. $t=1$ when $w_j$ is a positive sample; $t=0$ otherwise. Next we take the derivative of $E$ with regard to the output vector of the word $w_j$,
\begin{equation}
\frac{\partial E}{\partial \mathbf{v}'_{w_j}}
= \frac{\partial E}{\partial {\mathbf{v}'_{w_j}}^T\mathbf{h}} \cdot \frac{\partial {\mathbf{v}'_{w_j}}^T\mathbf{h}}{\partial \mathbf{v}'_{w_j}}
= \left(\sigma({\mathbf{v}'_{w_j}}^T\mathbf{h}) - t_j\right)\mathbf{h}
\end{equation}
which results in the following update equation for its output vector:
\begin{equation}
{\mathbf{v}'_{w_j}}^\text{(new)} = {\mathbf{v}'_{w_j}}^\text{(old)} - \eta\left(\sigma({\mathbf{v}'_{w_j}}^T\mathbf{h}) - t_j\right)\mathbf{h}
\end{equation}
which only needs to be applied to $w_j \in \{w_O\} \cup \mathcal{W}_{\text{neg}}$ instead of every word in the vocabulary. This shows why we may save a significant amount of computational effort per iteration.

The intuitive understanding of the above update equation should be the same as that of (\ref{eq:cbow-update-wij}). This equation can be used for both CBOW and the skip-gram model. For the skip-gram model, we apply this equation for one context word at a time.

To backpropagate the error to the hidden layer and thus update the input vectors of words, we need to take the derivative of $E$ with regard to the hidden layer's output, obtaining
\begin{eqnarray}
\frac{\partial E}{\partial \mathbf{h}}
&=& \sum_{w_j \in \{w_O\} \cup \mathcal{W}_{\text{neg}}}\frac{\partial E}{\partial {\mathbf{v}'_{w_j}}^T\mathbf{h}} \cdot \frac{\partial {\mathbf{v}'_{w_j}}^T\mathbf{h}}{\partial \mathbf{h}} \\
&=& \sum_{w_j \in \{w_O\} \cup \mathcal{W}_{\text{neg}}}\left(\sigma({\mathbf{v}'_{w_j}}^T\mathbf{h}) - t_j\right)\mathbf{v}'_{w_j} := \text{EH}
\end{eqnarray}
By plugging $\text{EH}$ into (\ref{eq:cbow-complex-update-wki}) we obtain the update equation for the input vectors of the CBOW model. For the skip-gram model, we need to calculate a $\text{EH}$ value for each word in the skip-gram context, and plug the sum of the $\text{EH}$ values into (\ref{eq:skipgram-update-wki}) to obtain the update equation for the input vector.

% I commented this section out as these are either not comprehensive nor insightful. The reader should go to other sources for in-depth understanding of why the model works.
%\section{Discussions}
%\subsection{Why softmax?}
%\label{sec:why-softmax}
%Using softmax can help us obtain a well-defined probabilistic (multinomial) distribution among words. The loss function for this model is cross-entropy. Theoretically we can also use squared sum as the error function, but then the efficiency performance optimization tricks developed for softmax will not apply.

%\subsection{Relationship between input vectors and output vectors?}
%Input vectors are taken from rows of the input$\rightarrow$hidden weight matrix, and output vectors are taken from columns of the hidden$\rightarrow$output weight matrix. By definition they are two different vector representations for words. 

%Based on the parameter update equations (\ref{eq:cbow-update-wki}) and (\ref{eq:cbow-update-wij}), intuitively the vectors for the same word should be close to each other. For hierarchical softmax, the output vectors have different meanings; but in negative sampling, one can experiment with forcing the two vectors to be the same and see what will happen.

%\subsection{Why stochastic gradient descent?}
%Stochastic gradient descent has been proven effective for learning large-scale training data with neural networks. However, one should pay attention to carefully tuning the learning rate and its descent rate.

\section*{Acknowledgement}
The author would like to thank Eytan Adar, Qiaozhu Mei, Jian Tang, Dragomir Radev,  Daniel Pressel, Thomas Dean, Sudeep Gandhe, Peter Lau, Luheng He, Tomas Mikolov, Hao Jiang, and Oded Shmueli for discussions on the topic and/or improving the writing of the note.

\bibliographystyle{apalike}
\bibliography{w2vexp}   

\newpage
\appendix
\section{Back Propagation Basics}
\label{sec:basics}

\subsection{Learning Algorithms for a Single Unit}
\begin{figure}[htbp]
\centering
\includegraphics[width=2.5in]{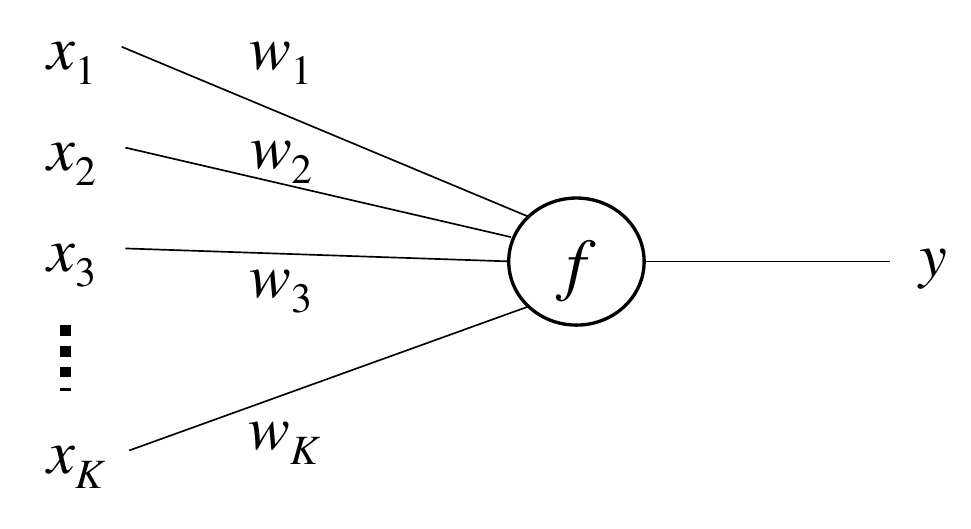}
\caption{An artificial neuron}
\label{fig:neuron}
\end{figure}

Figure~\ref{fig:neuron} shows an artificial neuron (unit). $\{x_1, \cdots, x_K\}$ are input values; $\{w_1, \cdots, w_K\}$ are weights; $y$ is a scalar output; and $f$ is the link function (also called activation/decision/transfer function).

The unit works in the following way:
\begin{equation}
y = f(u),
\end{equation}
where $u$ is a scalar number, which is the net input (or ``new input'') of the neuron. $u$ is defined as
\begin{equation}
u = \sum_{i=0}^K{w_i x_i}.
\end{equation}
Using vector notation, we can write
\begin{equation}
u = \mathbf{w}^T\mathbf{x}
\end{equation}

Note that here we ignore the bias term in $u$. To include a bias term, one can simply add an input dimension (e.g., $x_0$) that is constant 1.

Apparently, different link functions result in distinct behaviors of the neuron. We discuss two example choices of link functions here.

The first example choice of $f(u)$ is the \textbf{unit step function} (aka \textbf{Heaviside step function}):
\begin{equation}
f(u) = \begin{cases}1 & \text{if }u > 0\\0 & \text{otherwise}\end{cases}
\end{equation}

A neuron with this link function is called a perceptron. The learning algorithm for a perceptron is the perceptron algorithm. Its update equation is defined as:
\begin{equation}
\mathbf{w}^\text{(new)} = \mathbf{w}^\text{(old)} - \eta\cdot (y - t)\cdot \mathbf{x}
\end{equation}
where $t$ is the label (gold standard) and $\eta$ is the learning rate $(\eta > 0)$. Note that a perceptron is a linear classifier, which means its description capacity can be very limited. If we want to fit more complex functions, we need to use a non-linear model.

The second example choice of $f(u)$ is the \textbf{logistic function} (a most common kind of \textbf{sigmoid function}), defined as
\begin{equation}
\sigma(u) = \frac{1}{1 + e^{-u}}
\end{equation}
The logistic function has two primary good properties: (1) the output $y$ is always between 0 and 1, and (2) unlike a unit step function, $\sigma(u)$ is smooth and differentiable, making the derivation of update equation very easy.

Note that $\sigma(u)$ also has the following two properties that can be very convenient and will be used in our subsequent derivations:
\begin{equation}
\sigma(-u) = 1 - \sigma(u)
\label{eq:sigma-1}
\end{equation}
\begin{equation}
\frac{d\sigma(u)}{du} = \sigma(u)\sigma(-u)
\label{eq:sigma-2}
\end{equation}

We use stochastic gradient descent as the learning algorithm of this model. In order to derive the update equation, we need to define the error function, i.e., the training objective. The following objective function seems to be convenient:
\begin{equation}
E = \frac{1}{2}(t - y)^2
\end{equation}

We take the derivative of $E$ with regard to $w_i$,
\begin{eqnarray}
\frac{\partial E}{\partial w_i}
& = & \frac{\partial E}{\partial y} \cdot \frac{\partial y}{\partial u} \cdot \frac{\partial u}{\partial w_i} \\
& = & (y-t) \cdot y(1-y) \cdot x_i
\end{eqnarray}
where $\frac{\partial y}{\partial u} = y(1-y)$ because $y=f(u)=\sigma(u)$, and (\ref{eq:sigma-1}) and (\ref{eq:sigma-2}). Once we have the derivative, we can apply stochastic gradient descent:
\begin{equation}
\mathbf{w}^\text{(new)} = \mathbf{w}^\text{(old)} - \eta\cdot (y - t) \cdot y(1-y) \cdot \mathbf{x}.
\end{equation}

\subsection{Back-propagation with Multi-Layer Network}

\begin{figure}[htbp]
\centering
\includegraphics[width=3in]{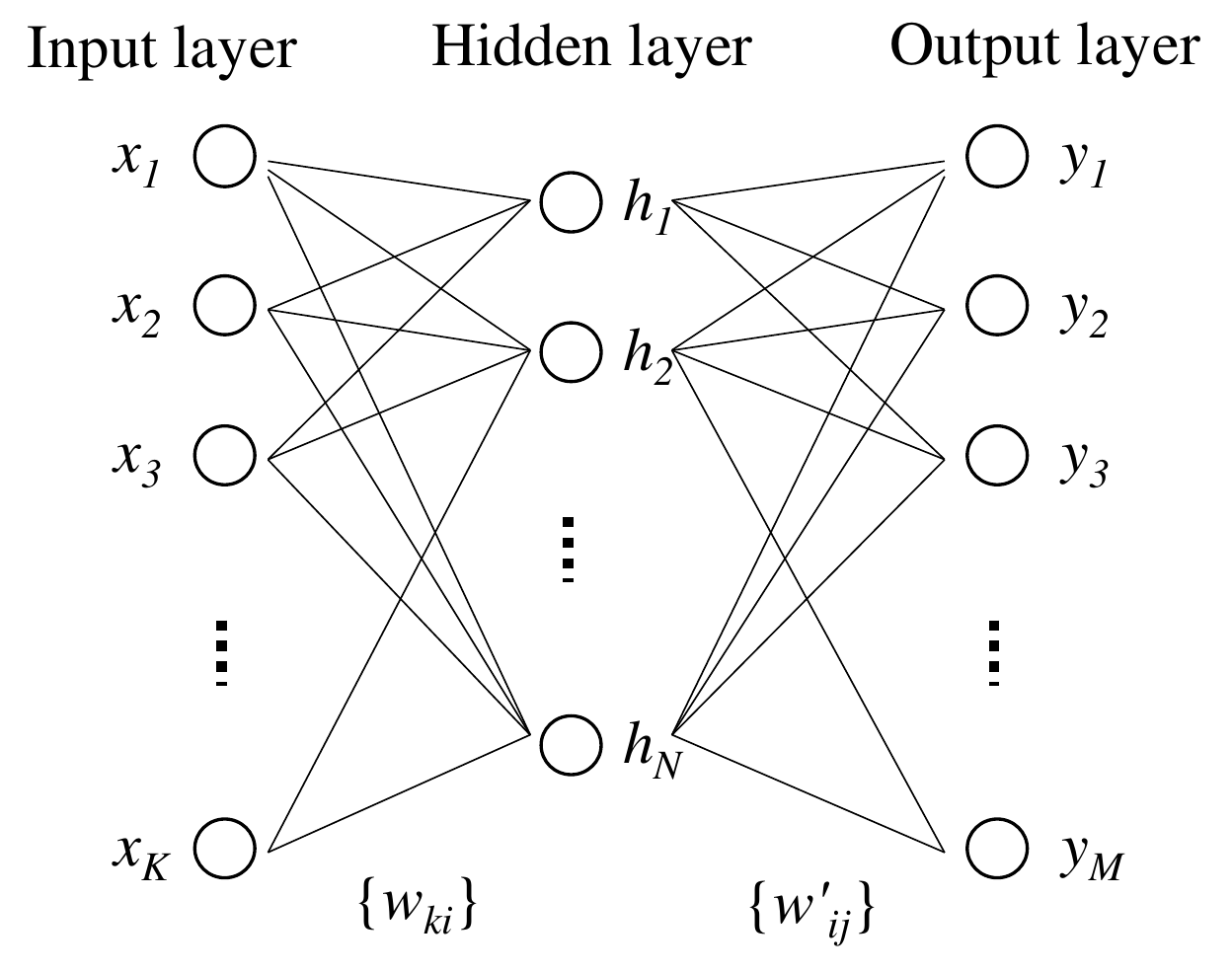}
\caption{A multi-layer neural network with one hidden layer}
\label{fig:multilayer}
\end{figure}

Figure~\ref{fig:multilayer} shows a multi-layer neural network with an input layer $\{x_k\} = \{x_1, \cdots, x_K\}$, a hidden layer $\{h_i\} = \{h_1, \cdots, h_N\}$, and an output layer $\{y_j\} = \{y_1, \cdots, y_M\}$. For clarity we use $k, i, j$ as the subscript for input, hidden, and output layer units respectively. We use $u_i$ and $u'_j$ to denote the net input of hidden layer units and output layer units respectively.

We want to derive the update equation for learning the weights $w_{ki}$ between the input and hidden layers, and $w'_{ij}$ between the hidden and output layers. We assume that all the computation units (i.e., units in the hidden layer and the output layer) use the logistic function $\sigma(u)$ as the link function. Therefore, for a unit $h_i$ in the hidden layer, its output is defined as 
\begin{equation}
h_i = \sigma(u_i) = \sigma\left(\sum_{k=1}^K w_{ki} x_k\right).
\end{equation}
Similarly, for a unit $y_j$ in the output layer, its output is defined as
\begin{equation}
y_j = \sigma(u'_j) = \sigma\left(\sum_{i=1}^N w'_{ij} h_i\right).
\end{equation}

We use the squared sum error function given by
\begin{equation}
E(\mathbf{x}, \mathbf{t}, \mathbf{W}, \mathbf{W'}) = \frac{1}{2}\sum_{j=1}^M(y_j - t_j)^2,
\end{equation}
where $\mathbf{W} = \{w_{ki}\}$, a $K\times N$ weight matrix (input-hidden), and $\mathbf{W'} = \{w'_{ij}\}$, a $N\times M$ weight matrix (hidden-output). $\mathbf{t} = \{t_1, \cdots, t_M\}$, a $M$-dimension vector, which is the gold-standard labels of output.

To obtain the update equations for $w_{ki}$ and $w'_{ij}$, we simply need to take the derivative of the error function $E$ with regard to the weights respectively. To make the derivation straightforward, we do start computing the derivative for the right-most layer (i.e., the output layer),
and then move left. For each layer, we split the computation into three steps, computing the derivative of the error with regard to the output, net input, and weight respectively. This process is shown below.

We start with the output layer. The first step is to compute the derivative of the error w.r.t. the output:
\begin{equation}
\frac{\partial E}{\partial y_j} = y_j - t_j.
\end{equation}
The second step is to compute the derivative of the error with regard to the net input of the output layer. Note that when taking derivatives with regard to something, we need to keep everything else fixed. Also note that this value is very important because it will be reused multiple times in subsequent computations. We denote it as $\text{EI}'_j$ for simplicity.
\begin{equation}
\frac{\partial E}{\partial u'_j} 
= \frac{\partial E}{\partial y_j}\cdot\frac{\partial y_j}{\partial u'_j}
= (y_j - t_j) \cdot y_j (1-y_j)
:= \text{EI}'_j
\label{eq:backprop1}
\end{equation}
The third step is to compute the derivative of the error with regard to the weight between the hidden layer and the output layer.
\begin{equation}
\frac{\partial E}{\partial w'_{ij}}
= \frac{\partial E}{\partial u'_j}  \cdot\frac{\partial u'_j}{\partial w'_{ij}}
= \text{EI}'_j \cdot h_i
\end{equation}
So far, we have obtained the update equation for weights between the hidden layer and the output layer.
\begin{eqnarray}
{w'_{ij}}^\text{(new)} 
&=& {w'_{ij}}^\text{(old)} - \eta \cdot \frac{\partial E}{\partial w'_{ij}} \\
&=& {w'_{ij}}^\text{(old)} - \eta \cdot \text{EI}'_j \cdot h_i.
\end{eqnarray}
where $\eta > 0$ is the learning rate.

We can repeat the same three steps to obtain the update equation for weights of the previous layer, which is essentially the idea of back propagation. 

We repeat the first step and compute the derivative of the error with regard to the output of the hidden layer. Note that the output of the hidden layer is related to all units in the output layer.
\begin{equation}
\frac{\partial E}{\partial h_i}
=\sum_{j=1}^M\frac{\partial E}{\partial u'_j}\frac{\partial u'_j}{\partial h_i}
=\sum_{j=1}^M \text{EI}'_j \cdot w'_{ij}.
\end{equation}
Then we repeat the second step above to compute the derivative of the error with regard to the net input of the hidden layer. This value is again very important, and we denote it as $\text{EI}_i$.
\begin{equation}
\frac{\partial E}{\partial u_i}
= \frac{\partial E}{\partial h_i} \cdot \frac{\partial h_i}{\partial u_i}
= \sum_{j=1}^M \text{EI}'_j \cdot w'_{ij} \cdot h_i(1-h_i)
:= \text{EI}_i
\label{eq:backprop2}
\end{equation}
Next we repeat the third step above to compute the derivative of the error with regard to the weights between the input layer and the hidden layer.
\begin{equation}
\frac{\partial E}{\partial w_{ki}}
= \frac{\partial E}{\partial u_i} \cdot \frac{\partial u_i}{\partial w_{ki}}
= \text{EI}_i \cdot x_k,
\end{equation}
Finally, we can obtain the update equation for weights between the input layer and the hidden layer.
\begin{equation}
{w_{ki}}^\text{(new)} = {w_{ki}}^\text{(old)} - \eta \cdot \text{EI}_i \cdot x_k.
\end{equation}

From the above example, we can see that the intermediate results ($\text{EI}'_j$) when computing the derivatives for one layer can be reused for the previous layer. Imagine there were another layer prior to the input layer, then $\text{EI}_i$ can also be reused to continue computing the chain of derivatives efficiently. Compare Equations (\ref{eq:backprop1}) and (\ref{eq:backprop2}), we may find that in (\ref{eq:backprop2}), the factor $\sum_{j=1}^M\text{EI}'_j w'_{ij}$ is just like the ``error'' of the hidden layer unit $h_i$. We may interpret this term as the error ``back-propagated'' from the next layer, and this propagation may go back further if the network has more hidden layers.

\section{wevi: Word Embedding Visual Inspector}
\label{sec:wevi}

\begin{figure}[htbp]
\centering
\includegraphics[width=6.5in]{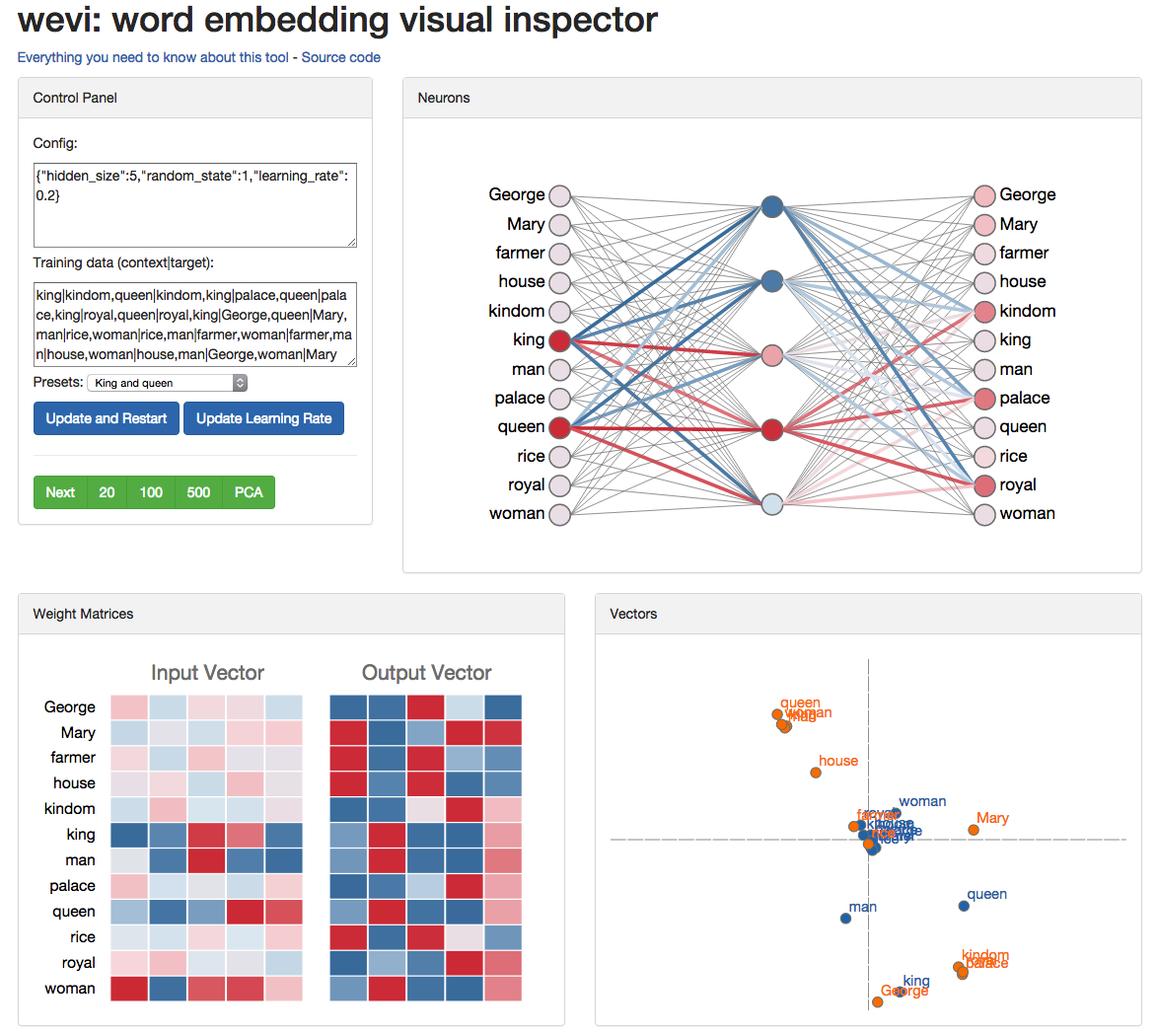}
\caption{wevi screenshot (\url{http://bit.ly/wevi-online})}
\label{fig:wevi}
\end{figure}

An interactive visual interface, wevi (word embedding visual inspector), is available online to demonstrate the working mechanism of the models described in this paper. See Figure~\ref{fig:wevi} for a screenshot of wevi. 

The demo allows the user to visually examine the movement of input vectors and output vectors as each training instance is consumed. The training process can be also run in batch mode (e.g., consuming 500 training instances in a row), which can reveal the emergence of patterns in the weight matrices and the corresponding word vectors. Principal component analysis (PCA) is employed to visualize the ``high''-dimensional vectors in a 2D scatter plot. The demo supports both CBOW and skip-gram models.

After training the model, the user can manually \textit{activate} one or multiple input-layer units, and inspect which hidden-layer units and output-layer units become active. The user can also customize training data, hidden layer size, and learning rate. Several preset training datasets are provided, which can generate different results that seem interesting, such as using a toy vocabulary to reproduce the famous word analogy: \textit{king} - \textit{queen} = \textit{man} - \textit{woman}.

It is hoped that by interacting with this demo one can quickly gain insights of the working mechanism of the model. The system is available at \url{http://bit.ly/wevi-online}. The source code is available at \url{http://github.com/ronxin/wevi}.

\end{document}